# Automated Multidisciplinary Design and Control of Hopping Robots for Exploration of Extreme Environments on the Moon and Mars

## Himangshu Kalita[a], Jekan Thangavelautham[b]*

[a] *Space and Terrestrial Robotics Exploration (SpaceTREx) Laboratory, Department of Aerospace and Mechanical Engineering, University of Arizona, 1130 N Mountain Ave, Tucson, Arizona 85721*, hkalita@email.arizona.edu
[b] *Space and Terrestrial Robotics Exploration (SpaceTREx) Laboratory, Department of Aerospace and Mechanical Engineering, University of Arizona, 1130 N Mountain Ave, Tucson, Arizona 85721*, jekan@email.arizona.edu
* Corresponding Author

## Abstract

The next frontier in solar system exploration will be missions targeting extreme and rugged environments such as caves, canyons, cliffs and crater rims of the Moon, Mars and icy moons. These environments are time capsules into early formation of the solar system and will provide vital clues of how our early solar system gave way to the current planets and moons. These sites will also provide vital clues to the past and present habitability of these environments. Current landers and rovers are unable to access these areas of high interest due to limitations in precision landing techniques, need for large and sophisticated science instruments and a mission assurance and operations culture where risks are minimized at all costs. Our past work has shown the advantages of using multiple spherical hopping robots called SphereX for exploring these extreme environments. Our previous work was based on performing exploration with a human-designed baseline design of a SphereX robot. However, the design of SphereX is a complex task that involves a large number of design variables and multiple engineering disciplines. In this work we propose to use Automated Multidisciplinary Design and Control Optimization (AMDCO) techniques to find near optimal design solutions in terms of mass, volume, power, and control for SphereX for different mission scenarios. The implementation of AMDCO for SphereX design is a complex process because of complexity of modelling and implementation, discontinuities in the design space, and wide range of time scales and exploration objectives. Moreover, the design of SphereX will depend on target environment (e.g. gravity, temperature, radiation and surface properties), coordination complexity with increased number of robots, expected distance of exploration and expected mission time length. We address these issues by using machine learning in the form of Genetic Algorithms integrated with gradient-based optimization techniques to search through the design space and find pareto optimal solutions for a given mission task. Using this technology, it is now possible to perform end to end automated preliminary design of planetary robots for surface exploration.
**Keywords:** (Multidisciplinary optimization, Genetic algorithms, Automated design)

## 1. Introduction

In the next few decades, we aspire to send human and robotic explorers to every corner of the solar system to perform orbital, surface and even subsurface exploration. These explorers will pave the way towards identifying the diverse surface environments, physical processes and structure of the planets and small bodies answering fundamental questions about the origins of the solar system, conditions to sustain life and prospects for resource utilization and off-world human settlement. Achieving this major exploration milestone remains technologically daunting but not impossible. An emerging target are the extreme environments of the Moon, Mars and icy moons, including caves, canyons, cliffs, skylights and craters (see Fig. 1). These are high-priority targets as outlined in the Planetary Science Decadal survey [1]. These environments are rich targets of origin studies, while caves offer natural shelter from radiation, harsh surface processes such as dust storms and are generally insulated by the varying high and low external temperatures. These conditions could harbour isolated, ancient ecosystems.

Exploration of these extreme and rugged environments remains out of reach from current planetary rovers and landers; however, the 2015 NASA Technology Roadmaps prioritizes the need for next-generation robotic and autonomous systems that can explore these extreme and rugged environment [2]. The challenges are three-fold and stem from current landing technology that requires wide-open spaces with no obstacles or landing hazards. A second challenge stems from current planetary vehicle architectures that house a growing variety of sophisticated science instruments. A third challenge has been the high standards of mission assurance expected. Due to the high costs and prestige for the nations involved, any form of exploratory risk that may reduce the life of the mission or result in damage to one or more subsystems is avoided. This is despite the potential science rewards from taking these exploratory risks. A credible solution is to develop an architecture that permits taking high exploratory risks that translates into high reward science but without compromising the rest of the mission.

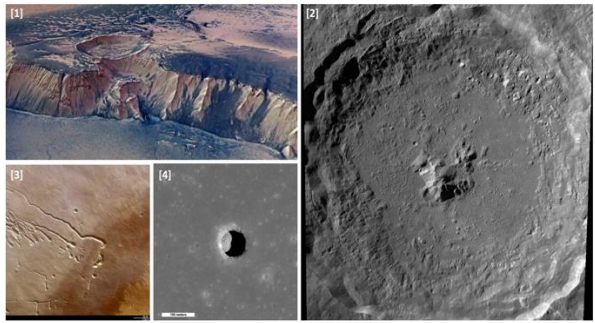

Fig. 1. Extreme environments of the Moon and Mars: (1) High cliffs surrounding Echus Chasma on Mars (nasa.gov), (2) Tycho crater on Moon (NASA/Goddard/Arizona State University), (3) Lava tubes on Pavonis Mons on Mars (ESA), and (4) Mare Tranquilitatis pit on Moon (NASA/GSFC/Arizona State University).

We present an architecture of small, low-cost, modular spherical robot called SphereX that is designed for exploring extreme environments like caves, lava tubes, pits and canyons in low-gravity environments like the Moon, Mars, icy moons and asteroids (see Fig. 2) [3-5,17]. It consists of a mobility system to perform optimal exploration of these target environments. It also consists of space-grade electronics like computer board for command and data handling, power board for power management and radio transceiver for communicating among multiple robots. Moreover, it also consists of a power system for power generation/storage, multiple UHF/S-band antennas and accommodates payloads in the rest of the volume. A large rover or lander may carry several of these SphereX robots that can be tactically deployed to explore and access rugged environments inaccessible by it.

However, the design of SphereX is a complex task that involves a large number of variables and multiple engineering disciplines. It is a highly coupled problem between multiple disciplines (see Fig. 2), and it must balance payload objectives against its overall size, mass, power and control which affects its cost and operation. Moreover, each subsystem has multiple candidate solutions, e.g. mobility can be achieved through hopping, rolling or wheels, power system can be designed through batteries that carries all the required power or can be generated on demand through fuel cells. Similarly, the selection of communication system and the avionics depends on numerous Commercially-Off-the-Shelf (COTS) options available, the thermal system can be designed through active, passive or a combination of both. As such finding optimal design solutions for SphereX to meet a defined mission requirement is of paramount importance. Currently, space systems are optimized manually through evaluation of each discipline independently. With this labour-intensive approach, although feasibility is achieved, there is no guarantee for achieving optimality of the overall system. Thus, space system design could benefit from the application of multidisciplinary design optimization (MDO).

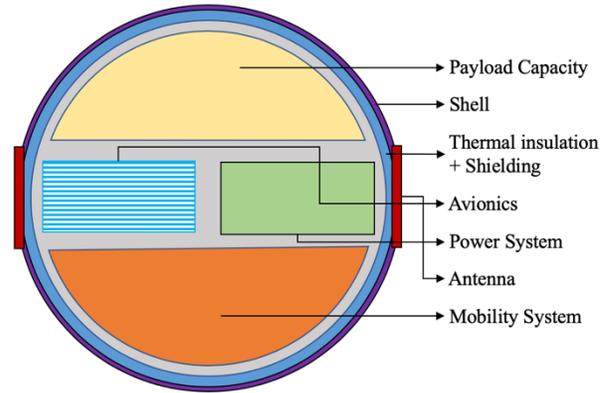

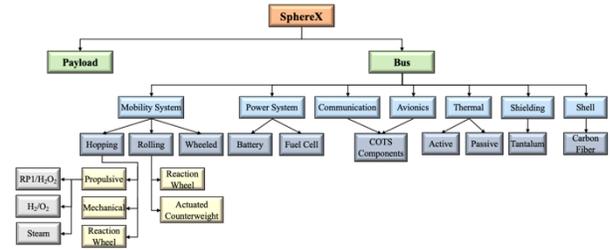

Fig. 2. (Top) SphereX architecture, (Bottom) Available options for each subsystem of SphereX.

However, complexity arises in an MDO approach due to complexity of modelling, complexity and discontinuity in objective cost function and design space, and wide range of time scales and mission requirements. Here, we approach this problem by using a hybrid optimization process where the search of the design space is performed with a GA based multi-objective optimizer at the system level to find the Pareto-optimal results while using gradient-based techniques at the discipline level. The methodology developed in this work uses Automated Multidisciplinary Design and Control Optimization (AMDCO) techniques to find near optimal design solutions in terms of mass, volume, power and control for SphereX for different mission scenarios. Using this technology, it is now possible to perform end to end automated preliminary design of planetary robots for extreme environment exploration.

For implementation perspective, the large number of disciplines of SphereX presents a significant challenge as they are coupled together. Fig. 3 shows the different disciplines of SphereX and how they are coupled together. The mission specifications and environment model affect design decisions of multiple disciplines of SphereX. For e.g. target distance affects the design of the mobility system, target mission time affects the power system and multiple number of robots introduces complexity in the communication system. Moreover, gravitational and surface properties models affect the design of the mobility system, radiation and temperature models affect the design of the thermal and shielding subsystems. Similarly, the mass and volume of each subsystem affect the design of the mobility system, and power requirement of each subsystem affect the design of the power system which in turn increase/decrease the mass and size of the power system affecting the mobility system. Furthermore, to increase payload volume, if we increase the size of the shell of the robot, its mass and inertia increases in the order of $\mathcal{O}(r^2)$, $r$ being the radius of the shell, thus affecting the mobility system. As such finding optimal design for each subsystem taking inter-subsystem dependencies into account is of paramount importance.

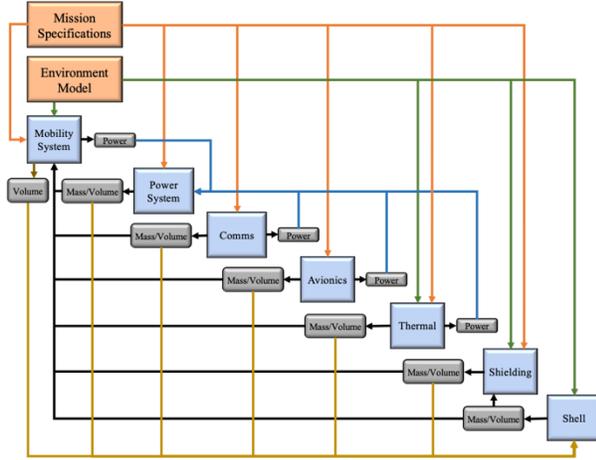

Fig. 3. Design structure matrix of relevant disciplines for SphereX.

## 2. Approach and methods

A multidisciplinary design and control optimization-based approach is proposed to explore the question of how to maximize the payload mass, volume and power budget while minimizing the total mass, volume and power of SphereX. The problem is approached by using a hybrid optimization process where the search of the design space is performed with a multi-objective optimizer at the system level to find the Pareto-optimal results while using gradient-based techniques at the discipline level as shown is Fig. 4. At the system level, the multi-objective optimization problem is formulated as Eq. (1).

$$\min \mathbb{F}_k(\mathbb{x}) \quad k = 1,2,\dots,K$$
$$s.t. \begin{cases} \mathbb{G}_l(\mathbb{x}) \le 0 & l = 1,2,\dots,L \\ \mathbb{H}_m(\mathbb{x}) = 0 & m = 1,2,\dots,M \\ \mathbb{x}_j^{(L)} \le \mathbb{x}_j \le \mathbb{x}_j^{(U)} & j = 1,2,\dots,J \end{cases} \quad (1)$$

The solution $\mathbb{x}$ is a vector of $J$ system level design variables: $\mathbb{x} = [\mathbb{x}_1, \mathbb{x}_2, \dots, \mathbb{x}_J]^T$. There are $K$ objective functions $\mathbb{F} = [\mathbb{F}_1, \mathbb{F}_2, \dots, \mathbb{F}_K]^T$. Associated with the problem are $L$ inequality constraints and $M$ equality constraints. The last set of $J$ constraints are the variable bounds, restricting each decision variable $\mathbb{x}_j$ to take a value within a lower $\mathbb{x}_j^{(L)}$ and an upper $\mathbb{x}_j^{(U)}$ bound.

Each of the subsystem discipline are modelled as a single-objective optimization problem and for each subsystem model $i = 1,2,\dots,N$, the problem is formulated as Eq. (2).

$$\min f_i(\mathbb{d}_i)$$
$$s.t. \begin{cases} g_{ip}(\mathbb{d}_i) \le 0 & p = 1,2,\dots,P_i \\ h_{iq}(\mathbb{d}_i) = 0 & q = 1,2,\dots,Q_i \\ \mathbb{d}_{ir}^{(L)} \le \mathbb{d}_{ir} \le \mathbb{d}_{ir}^{(U)} & r = 1,2,\dots,R_i \end{cases} \quad (2)$$

For each subsystem discipline $i$, $\mathbb{d}_i$ is a vector of $R_i$ discipline level design variables: $\mathbb{d}_i = [\mathbb{d}_{i1}, \mathbb{d}_{i2}, \dots, \mathbb{d}_{iR_i}]^T$. Each discipline has $P_i$ inequality constraints, $Q_i$ equality constraints, and $R_i$ constraints that restricts the design variables within the upper and lower bounds. The vector $\mathbb{y}$ represents the specifications of the COTS components selected for each subsystem based on the inventory list and the system level design variables $\mathbb{x}$. Moreover, $\mathbb{c}_{ij}, \forall i \in \{1,\dots,N\}, \forall j \in \{1,\dots,N\}, i \ne j$ are the coupling functions calculated by discipline $i$ and input to discipline $j$.

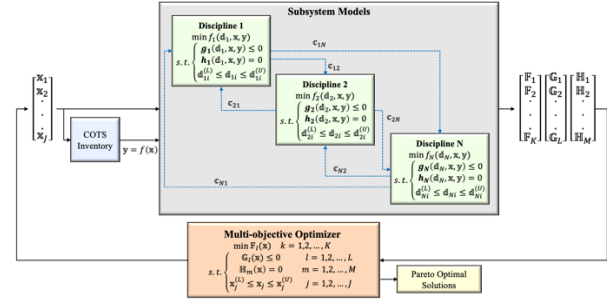

Fig. 4. Hybrid optimization approach for multidisciplinary optimization.

The approach is introduced through the Automated Multidisciplinary Design and Control Optimization (AMDCO) software tool as shown in Fig. 5. As seen here, the software has 4 primary blocks: a user input interface, a system designer base, a path planner base, and an output user interface.

The designer defines the high-level mission specifications and environment model in the user interface. The mission specifications include parameters such as 1. Target distance, 2. Target mission length, and 3. Number of robots deployed, and the environment model include 1. Gravitational model, 2. Radiation model, 3. Temperature model, and 4. Surface properties. These parameters are then sent to the system designer, which then solves each aspect of all the subsystems. The system designer continuously interacts with each subsystem models, COTS inventory for electronics components, and controllers to find Pareto-optimal solutions for the overall system. The output interface then plots the entire pareto-optimal solutions. The user can select any solution in the pareto front and analyse the respective autonomously assembled 3D model of the entire system, performance analysis of the mobility controller and details of each subsystem. Moreover, the user can input 3D maps of target environment for exploration (e.g. caves, lava tubes, pits, planetary surface etc.) and select a solution from the pareto front to perform detailed path planning analysis. Two options are incorporated in the software where the user can perform path planning with a single robot or multiple robots in the target environment. The approach extends the traditional system design by merging optimal geometric design with control design for SphereX.

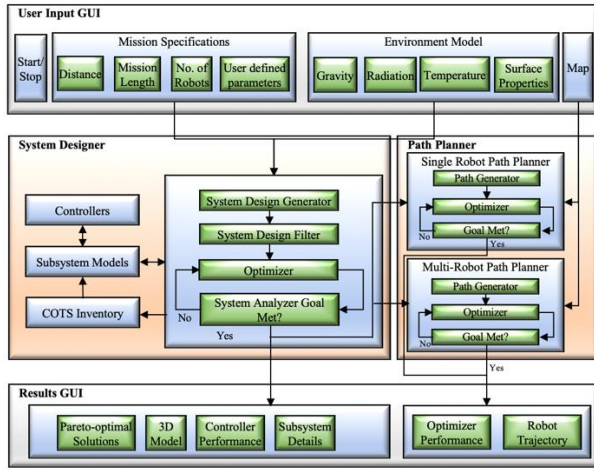

Fig. 5. Software architecture for Automated Multidisciplinary Design and Control Optimization (AMDCO) software to provide end-to-end design framework for SphereX.

In order to find optimal design solutions of SphereX, multiple subsystem discipline level single-objective optimization problems, and a system level multi-objective optimization problem are to be solved simultaneously. This section provides the methods used in this research to solve the single-objective and multi-objective optimization problems.

## 2.1 Single objective optimization

As discussed above, the subsystem discipline models are modelled as single-objective optimization problem with one objective function, $P$ inequality constraints, $Q$ equality constraints and $R$ side constraints. The $R$ side constraints are converted into $2R$ inequality constraints such that there are $S = P + 2R$ inequality constraints, thus the problem is modelled as a nonlinear optimization problem (NLP) of the form shown in Eq. (3).

$$\min_{d \in \mathbb{R}} f(d)$$
$$subject\ to \begin{cases} g(d) \leq 0 \\ h(d) = 0 \end{cases} \quad (3)$$

Where, $f: \mathbb{R}^R \to \mathbb{R}$ is the objective function, the function $g: \mathbb{R}^R \to \mathbb{R}^S$ and $h: \mathbb{R}^R \to \mathbb{R}^Q$ are the inequality and equality constraints. Moreover, a scalar-valued function $\mathcal{L}: \mathbb{R}^{R \times S \times Q} \to \mathbb{R}$ is defined called the Lagrangian function of the NLP by Eq. (4)

$$\mathcal{L}(d, \lambda, \mu) = f(d) + \lambda^T h(d) + \mu^T g(d) \quad (4)$$

The vectors $\lambda \in \mathbb{R}^Q$ and $\mu \in \mathbb{R}^S$ are the Lagrange multiplier vectors. Given a vector $d$, the set of active constraints at $d$ consists of the inequality constraints, if any, satisfied as equalities at $d$ denoted by $g^{\blacksquare}(d)$. For $d^* \in \mathbb{R}^R$ to be an isolated local minimum of the NLP, the Karush-Kuhn-Tucker (KKT) conditions should apply [6]. To solve the NLP problem, The Sequential Quadratic Programming (SQP) method is used which is an iterative method in which, at a current iterate $d^k$, the step to the next iterate is obtained through information given by solving a quadratic subproblem as shown in Eq. (5).

$$\min_{d \in \mathbb{R}} \nabla f(d^k)^T d(d) + \frac{1}{2} d(d)^T H f(d^k) d(d)$$
$$s.t. \begin{cases} g(d^k) + \nabla g(d^k)^T d(d) \leq 0 \\ h(d^k) + \nabla h(d^k)^T d(d) = 0 \end{cases} \quad (5)$$

where, $d(d) = d - d^k$, $\nabla f$ is the gradient and $Hf$ is the hessian. The QP is related to the local quadratic model of the Lagrangian $\mathcal{L}$ as the objective function which leads to the QP subproblem as shown in Eq. (6).

$$\min_{d \in \mathbb{R}} \nabla \mathcal{L}^T d(d) + \frac{1}{2} d(d)^T H \mathcal{L} d(d)$$
$$s.t. \begin{cases} g(d^k) + \nabla g(d^k)^T d(d) \leq 0 \\ h(d^k) + \nabla h(d^k)^T d(d) = 0 \end{cases} \quad (6)$$

The local convergence of the SQP method follows from the application of Newton's method to the nonlinear system given by the KKT conditions as shown in Eq. (7).

$$\Psi(d, \lambda, \mu) = \begin{bmatrix} \nabla \mathcal{L}(d, \lambda, \mu) \\ h(d) \\ g^{\blacksquare}(d) \end{bmatrix} = 0 \quad (7)$$

The Jacobian of the nonlinear system is given by Eq. (8).

$$J(d, \lambda, \mu) = \begin{bmatrix} H\mathcal{L}(d, \lambda, \mu) & \nabla h(d) & \nabla g^{\blacksquare}(d) \\ \nabla h(d) & 0 & 0 \\ \nabla g^{\blacksquare}(d) & 0 & 0 \end{bmatrix} \quad (8)$$

Therefore, the Newton iteration is given by Eq. (9).

$$\begin{aligned} d^{k+1} &= d^k + s_d \\ \lambda^{k+1} &= \lambda^k + s_\lambda \\ \mu^{k+1} &= \mu^k + s_\mu \end{aligned} \quad (9)$$

where, $s = (s_d, s_\lambda, s_\mu)$ is the solution of Eq. (10).

$$J(d^k, \lambda^k, \mu^k)s = -\Psi(d^k, \lambda^k, \mu^k) \quad (10)$$

## 2.2 Multi-objective optimization

The system level model is modelled as a multi-objective optimization problem with $K$ objective functions, $L$ inequality constraints, $M$ equality constraints and $J$ side constraints. The $J$ side constraints are converted into $2J$ inequality constraints such that there are $T = L + 2J$ inequality constraints, thus the problem is modelled as in Eq. (11).

$$\min \mathbb{F}_k(\mathbf{x}) \quad k = 1,2,\dots,K$$

$$s.t. \begin{cases} \mathbb{G}_t(\mathbf{x}) \leq 0 & t = 1,2,\dots,T \\ \mathbb{H}_m(\mathbf{x}) = 0 & m = 1,2,\dots,M \end{cases} \quad (11)$$

The constraints divide the search space into two divisions – feasible and infeasible regions. The constraints are handled by using the Penalty Function approach. For each solution $\mathbf{x}^{(i)}$, the constraint violation for the inequality constraints $\mathbb{G}_t(\mathbf{x}^{(i)})$ for $t = 1,2,\dots,T$ are calculated as in Eq. (12).

$$w_t(\mathbf{x}^{(i)}) = \begin{cases} |\mathbb{G}_t(\mathbf{x}^{(i)})|, & if \ \mathbb{G}_t(\mathbf{x}^{(i)}) > 0 \\ 0 & otherwise \end{cases} \quad (12)$$

The constraint violation for the equality constraints $\mathbb{H}_m(\mathbf{x})$ for $m = 1,2,\dots,M$ are calculated as in Eq. (13).

$$w_m(\mathbf{x}^{(i)}) = |\mathbb{H}_m(\mathbf{x}^{(i)})| \quad (13)$$

Thereafter, all constraint violations are added together to get the overall constraint violation as in Eq. (14).

$$\Omega(\mathbf{x}^{(i)}) = \sum_{t=1}^{T} w_t(\mathbf{x}^{(i)}) + \sum_{m=1}^{M} w_m(\mathbf{x}^{(i)}) \quad (14)$$

This constraint violation is then multiplied with a penalty parameter $\mathcal{P}_k$ and then the product is added to each of the objective function values as in Eq. (15).

$$\mathcal{J}_k(\mathbf{x}^{(i)}) = \mathbb{F}_k(\mathbf{x}^{(i)}) + \mathcal{P}_k\Omega(\mathbf{x}^{(i)}) \quad (15)$$

The cost function $\mathcal{J}_k$ takes into account the constraint violations. For a feasible solution the corresponding $\Omega$ term is zero and $\mathcal{J}_k$ becomes equal to the original objective function $\mathbb{F}_k$. However, for an infeasible solution, $\mathcal{J}_k > \mathbb{F}_k$, thereby adding a penalty corresponding to the total constraint violation. The resulting outcome of the multi-objective optimization process is a set of optimal solutions with a varying degree of objective values called the Pareto-optimal solutions. To solve multi-objective optimization problem, there are a few classical methods like 'Weighted Sum Method', 'ε-Constraint Method', 'Weighted Metric Method', 'Benson's Method', 'Value Function Method', and 'Goal Programming Method'. All these classical methods use a single solution update in every iteration and mainly use a deterministic transition rule, however, in case of evolutionary algorithms (EA), a population of solutions is processed in every iteration (or generation). This feature alone gives an EA a tremendous advantage for its use in solving multi-objective optimization problems (MOOPs) [7].

For this research, a real-parameter elitist non-dominated sorting genetic algorithm (NSGA-II) is used to find the pareto optimal solutions. Initially a random parent population of the design variables $P_0$ is created of size $N_P$. For each individual, the values of each cost functions are calculated, and the population is sorted based on nondomination and each solution is assigned a rank ($r$) equal to its nondomination level ($\mathcal{F}$) and a crowding distance ($d$). Since the problem is formulated as a minimization problem, the vector $\mathbf{x}^{(1)}$ is partially less than another vector $\mathbf{x}^{(2)}$, $(\mathbf{x}^{(1)} \prec \mathbf{x}^{(2)})$, when no value of $\mathbf{x}^{(2)}$ is less than $\mathbf{x}^{(1)}$ and at least one value of $\mathbf{x}^{(2)}$ is strictly greater than $\mathbf{x}^{(1)}$. If $\mathbf{x}^{(1)}$ is partially less than $\mathbf{x}^{(2)}$, the solution $\mathbf{x}^{(1)}$ dominates $\mathbf{x}^{(2)}$ [7]. Any member of such vectors which is not dominated by any other member is said to be nondominated. To get an estimate of the density of solutions surrounding a particular solution in a nondomination level, a quantity called crowding distance that serves as an estimate of the perimeter of the cuboid formed by using the nearest neighbours as vertices is calculated. For computing the crowding distance, first the number of solutions in $\mathcal{F}$ is calculated as $l = |\mathcal{F}|$, and for each $i$ in the set $d_i = 0$ is assigned. Next for each objective function $k$, the set is sorted in worst order of $\mathcal{J}_k$ and the sorted indices are stored in a vector: $I^k = sort(\mathcal{J}_k, >)$. Nest for each $k$, a large distance is assigned to the boundary solutions, $d_{I_1^k} = d_{I_l^k} = \infty$, and for all other solutions $j = 2 \ to \ (l-1)$, the distance is shown by Eq. (16).

$$d_{I_j^k} = d_{I_j^k} + \frac{\mathcal{J}_k^{(I_{j+1}^k)} - \mathcal{J}_k^{(I_{j-1}^k)}}{\mathcal{J}_k^{max} - \mathcal{J}_k^{min}} \quad (16)$$

The index $I_j$ denotes the solution index of the j-th member in the sorted list. The calculation is continued with other objective functions and the overall crowding distance value ($d$) is calculated as the sum of individual distance values corresponding to each objective. With the non-domination rank and crowding distance of each individual determined, the crowded tournament selection operator is used to select individuals for crossover [7]. The selection operator compares two solutions and returns the winner of the tournament. A solution $i$ wins a tournament with another solution $j$ if solution $i$ has a better rank, that is, $r_i < r_j$. If they have the same rank, solution $i$ wins if it has a better crowding distance, that is, $r_i = r_j$ and $d_i > d_j$. Next crossover, and mutation operators are used to create an offspring population $Q_0$ of size $N_Q$. For crossover, a blend crossover (BLX-$\alpha$) operator is used [8]. For two parent solutions $\mathbf{x}_i^{(1,t)}$ and $\mathbf{x}_i^{(2,t)}$ (assuming $\mathbf{x}_i^{(1,t)} < \mathbf{x}_i^{(2,t)}$) in generation ($t$), the BLX-$\alpha$ randomly picks a solution in the range $[\mathbf{x}_i^{(1,t)} - \alpha(\mathbf{x}_i^{(2,t)} - \mathbf{x}_i^{(1,t)}), \mathbf{x}_i^{(2,t)} + \alpha(\mathbf{x}_i^{(2,t)} - \mathbf{x}_i^{(1,t)})]$. Thus if $u_i$ is a random number between 0 and 1, the following is an offspring:

$$\mathbf{x}_i^{(1,t+1)} = (1 - \gamma_i)\mathbf{x}_i^{(1,t)} + \gamma_i \mathbf{x}_i^{(2,t)} \quad (17)$$

where, $\gamma_i = (1 + 2\alpha)u_i - \alpha$, which is uniformly distributed for a fixed value of $\alpha$. If $\alpha = 0$, this crossover creates a random solution in the range $(\mathbf{x}_i^{(1,t)}, \mathbf{x}_i^{(2,t)})$. In a number of test problems, the investigators have reported that BLX-0.5 (with $\alpha = 0.5$) performs better than with any other $\alpha$ value. Moreover, the location of the offspring depends on the difference in parent solutions. Equation (17) can be rewritten as:

$$x_i^{(1,t+1)} - x_i^{(1,t)} = \gamma_i\left(x_i^{(2,t)} - x_i^{(1,t)}\right) \quad (18)$$

If the difference between the parent solutions is small, the difference between the offspring and parent solution is also small. This property allows to constitute an adaptive search. If the diversity in the parent population is large, an offspring population with a large diversity is expected, and vice versa. Thus, this operator allows the searching of the entire space early on and also allow to maintain a focused search when the population tends to converge in some region in the search space. During mutation, a non-uniform mutation operator is used, where the probability of creating a solution closer to the parent is more than the probability of creating one away from it as shown by Eq. (19) [9]. However, as the generation ($t$) proceed, this probability of creating solutions closer to the parent gets higher and higher.

$$x_i^{(t+1)} = x_i^{(t)} + \tau\left(x_i^{(U)} - x_i^{(L)}\right)\left(1 - u_i^{\left(1-\frac{t}{t_{max}}\right)^b}\right) \quad (19)$$

Here, $\tau$ takes a Boolean value, -1 or 1, each with a probability of 0.5. The parameter $u_i$ is a random number between 0 and 1, $t_{max}$ is the maximum number of allowed generations, and $b$ is a user defined parameter. In this way, from early on the above mutation operator acts like a uniform distribution, while in later generations it acts like Dirac's function, thus allowing a focused search. For variables that have integer constraints, it is rounded off to the nearest integer after crossover and mutation. Since elitism is introduced by comparing current population with previously found best nondominated solutions, the procedure is different after the initial generation [7]. For the $t^{th}$ generation, first a combined population $R_t = P_t \cup Q_t$ is formed. The population $R_t$ is of size $N_P + N_Q$. Then, the population is sorted according to nondomination. Since all previous and current population members are included in $R_t$, elitism is ensured. Now, solutions belonging to the best nondominated set $\mathcal{F}_1$ are emphasized more than any other solution in the combined population. If the size of $\mathcal{F}_1$ is smaller than $N_P$, we choose all members of the set $\mathcal{F}_1$ for the new population $P_{t+1}$. The remaining members of the population $P_{t+1}$ are chosen from subsequent nondominated fronts in the order of their ranking. Thus, solutions from the set $\mathcal{F}_2$ are chosen next, followed by solutions from the set $\mathcal{F}_3$, and so on. This procedure is continued until no more sets can be accommodated. When the set $\mathcal{F}_l$ is the last nondominated set beyond which no other set can be accommodated, the count of solutions in all sets from $\mathcal{F}_1$ to $\mathcal{F}_l$ would be larger than the population size. To choose exactly $N_P$ population members, we sort the solutions of the last front $\mathcal{F}_l$ using the crowding distance operator in descending order and choose the best solutions needed to fill all population slots as shown in Fig. 6. The new population $P_{t+1}$ of size

$N_P$ is now used for selection, crossover, and mutation to create a new population $Q_{t+1}$ of size $N_Q$.

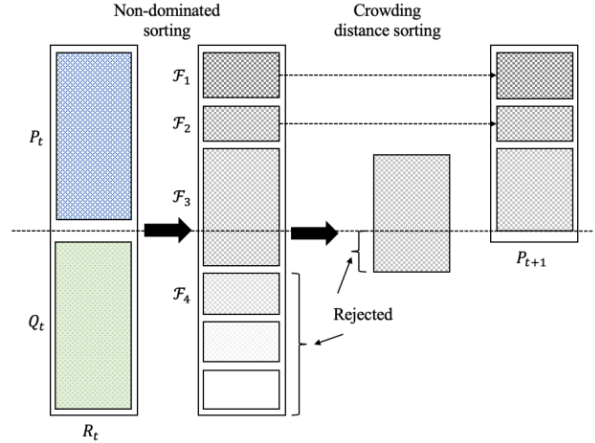

Fig. 6. Schematic of the NSGA-II procedure.

## 3. Environment and Subsystem Models

The ambient environmental factors present on the lunar and Martian surface pose some of the most difficult challenges for the success of long-term robotic exploration. These factors include dangerous radiation levels and high range of temperatures that can pose a variety of complications like thermal expansion and contraction, bit flips, and electrical leakage. Moreover, the dynamics and efficiency of the robot is dependent on the gravity and the surface interaction parameters. As such, the design of the robot should take these factors in account. Detailed models of the environmental factors: a) Temperature model, b) Radiation model, c) Gravitational model, and d) Surface interaction model for the surface of the Moon and Mars are developed based on literature review [10-16].

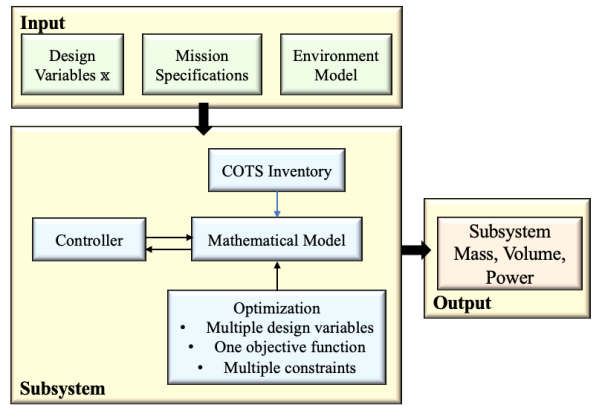

Fig. 7. Graphical representation of each subsystem discipline models.

Mathematical models for each subsystem disciplines of SphereX are also developed for this research. The modelled subsystems are mobility system, power system, thermal system, shielding, communication system, avionics and shell. Moreover, for the mobility subsystem, multiple controllers are developed that interacts with the mathematical model during each iteration of the optimization process. Each subsystem is defined by multiple design variables, one objective function and multiple equality, inequality and side constraints. The mathematical model is used for iteration to find the optimal design variables and the mass, volume and power requirements for each subsystem are calculated as shown in Fig. 7. The mass, volume and power requirements for each subsystem will then be used for the system level optimization process.

## 4. System Level Optimization

With all the subsystem models and their respective optimization models defined, this section defines the system level optimization model. The objective of our MDO approach is to find the optimum mass and radius of the robot (SphereX) that accommodates the maximum payload in terms of mass, volume and power based on predefined mission specifications. The problem is formulated as a multi-objective optimization problem (MOOP) with 12 design variables $\mathbb{x} = [m, r, P, ms_{ID}, sd_1, sd_2, ps_{ID}, c_{ID}, p_{ID}, b_{ID}, t_{ID}, a_{ID}]$, 4 objective functions and 5+ constraints, where $m$ and $r$ are the mass and radius of the robot, $P$ is the power demand, $ms_{ID}$ defines the type, and $sd_1$ and $sd_2$ the subtype of the mobility system, $ps_{ID}$ defines the type of power system and the COTS IDs are defined as main computer ($c_{ID}$), power management board ($p_{ID}$), battery ($b_{ID}$), radio transceiver board ($t_{ID}$), and attitude controller board ($a_{ID}$). Based on the values of mass $m$, radius $r$, and power demand $P$ with bounds $m^{(b)} = [m^{(L)} \quad m^{(U)}]$, $r^{(b)} = [r^{(L)} \quad r^{(U)}]$, and $P^{(b)} = [P^{(L)} \quad P^{(U)}]$ it is normalized between $[0 \quad 1]$ as shown in Eq. (20).

$$\underline{m} = \frac{m - m^{(L)}}{m^{(U)} - m^{(L)}}, \quad \underline{r} = \frac{r - r^{(L)}}{r^{(U)} - r^{(L)}},$$
$$\underline{P} = \frac{P - P^{(L)}}{P^{(U)} - P^{(L)}} \tag{20}$$

The first objective is then defined as $\mathbb{F}_1(\mathbb{x}) = \alpha_1\underline{m} + \alpha_2\underline{r}$. Based on the design variable, the mass, volume and power of each subsystem is calculated and the mass and volume of the payload is calculated as $m_{pay} = m - m_{sys}$, $\mathcal{V}_{pay} = \mathcal{V} - \mathcal{V}_{sys}$, and $P_{pay} = P - P_{sys}$, where $\mathcal{V} = 4\pi r^3/3$. The payload mass, volume and power ratio are then calculated as $m_r = m_{pay}/m$, $\mathcal{V}_r = \mathcal{V}_{pay}/\mathcal{V}$, and $P_r = P_{pay}/P$. The second objective function is then defined as $\mathbb{F}_2(\mathbb{x}) = 1 - (\alpha_3 m_r + \alpha_4 \mathcal{V}_r)$. $m_{sys}$, $\mathcal{V}_{sys}$, and $P_{sys}$ are the total mass, volume and power of all the subsystems described in section 6, and $\alpha_1$, $\alpha_2$, $\alpha_3$, and $\alpha_4$ are weights. The third and fourth objective functions are defined as $\mathbb{F}_3(\mathbb{x}) = \underline{P}$ and $\mathbb{F}_4(\mathbb{x}) = 1 - P_r$. Three constraints are added to the optimization problem. The first three constraints are $m_r > 0$, $\mathcal{V}_r > 0$ and $P_r > 0$. The fourth constraint is $Index = 1$. The fifth constraint is that the bandwidth of the transceiver selected lies within the resonating frequency of the antenna designed. Finally, other constraints can be added based on other user defined parameters, (e.g. the clock frequency of the computer selected is greater than a user-defined desired clock frequency, storage capacity of the computer selected is greater than a user-defined value etc.). The optimization problem is then mathematically formulated as Eq. (21).

$$\min \mathbb{F}_1(\mathbb{x}) = \alpha_1\underline{m} + \alpha_2\underline{r}$$
$$\min \mathbb{F}_2(\mathbb{x}) = 1 - (\alpha_3 m_r + \alpha_4\mathcal{V}_r)$$
$$\min \mathbb{F}_3(\mathbb{x}) = \underline{P}$$
$$\min \mathbb{F}_4(\mathbb{x}) = 1 - P_r$$
$$s.t. \begin{cases} \mathbb{G}_1(\mathbb{x}) \equiv m_r > 0 \\ \mathbb{G}_2(\mathbb{x}) \equiv \mathcal{V}_r > 0 \\ \mathbb{G}_3(\mathbb{x}) \equiv P_r > 0 \\ \mathbb{G}_4(\mathbb{x}) \equiv Index = 1 \\ \mathbb{G}_5(\mathbb{x}) \equiv f_{trans}^{(L)} + \frac{BW}{2} \leq f_{r(ant)} \leq f_{trans}^{(U)} - \frac{BW}{2} \end{cases} \tag{21}$$

Since our problem is a constrained multi-objective problem, the search space is divided into two regions: feasible and infeasible regions. Hence, all pareto optimal solutions must also lie in the feasible region. The penalty function approach was used to handle the constraints within the objective functions as discussed in Section 5. For each solution $\mathbb{x}^{(i)}$, the constraint violation for each constraint are calculated and then added together to get the overall constraint violation $\Omega(\mathbb{x}^{(i)})$. This constraint violation is then multiplied with a penalty parameter $\mathcal{P}$ and the product is added to each of the objective functions. Thus, the constrained multi-objective optimization problem is converted into an unconstrained multi-objective optimization problem with the 4 cost functions defined as Eq. (22).

$$\mathcal{J}_k(\mathbb{x}) = \mathbb{F}_k(\mathbb{x}) + \mathcal{P}\Omega(\mathbb{x}), \quad k = 1,2,3,4 \tag{22}$$

With the 4 cost functions defined, an elitist non-dominated sorting genetic algorithm (NSGA-II) is used to find the pareto optimal solutions as discussed in Section 2. For creating the initially random parent population $P_0$, the values of $m$, $r$ and $P$ are chosen with a uniform distribution $m = \mathcal{U}(m^{(U)}, m^{(L)})$, $r = \mathcal{U}(r^{(U)}, r^{(L)})$, and $P = \mathcal{U}(P^{(U)}, P^{(L)})$, the integer values of $ms_{ID}, sd_1, sd_2, ps_{ID}$ are chosen at random from the available options, and the integer COTS IDs are chosen at random from the COTS inventory.

## 5. Results and Discussion

This section provides the results of simulations performed for different mission scenarios. The simulation results are presented in the form of pareto optimal design solutions for two exploration missions 1) Surface exploration mission on Mare Tranquilitatis, and

2) Subsurface exploration mission of Mare Tranquilitatis pit on the surface of the Moon. The mission specifications were to explore 1000 meters over a mission lifetime of 5 hours and 3000 meters over a mission lifetime of 15 hours respectively. Along with the pareto optimal solutions, the history of selection of the mobility and power system is presented that showed the selection of optimal mobility and power system for different mission scenarios. To better understand the selection probability of the mobility and power system, a comparative analysis is presented for all combinations of propulsive mobility system and power system for varying mission exploration requirements.

## 5.1 Surface exploration on Mare Tranquilitatis

The first simulation was run to perform surface exploration on Mare Tranquilitatis. Mare Tranquilitatis is a lunar mare that sits within the Tranquilitatis basin on the Moon at 8.5°N 31.4°E, which was also the landing site for the first manned landing on the Moon (Apollo-11) on July 20, 1969. The mission target is to explore 1000m around the Apollo-11 landing site in 5 hours. The environmental conditions used for the simulations were gravity $g = 1.62 \text{m/s}^2$, ambient temperature $T_a = 340\text{K}$, radiation dose rate $I_0 = 100\text{rad/yr}$, and soil properties of Lunar soil.

Fig. 8 shows the mass, volume, and power budget of the robot for the pareto-optimal solutions found. It can be seen that the minimum and maximum values of mass, volume and power available for the payload are $m_{pay}^{(L)} = 0.20$, $m_{pay}^{(U)} = 4.56 \text{ kg}$, $\mathcal{V}_{pay}^{(L)} = 0.00022$, $\mathcal{V}_{pay}^{(U)} = 0.0109 \text{ m}^3$, and $P_{pay}^{(L)} = 0.001$, $P_{pay}^{(U)} = 22.81 \text{ W}$. The average values of the mass, volume and power available for the payload are $\bar{m}_{pay} = 2.27 \text{ kg}$, $\bar{\mathcal{V}}_{pay} = 0.0056 \text{ m}^3$, and $\bar{P}_{pay} = 9.04 \text{ W}$ over the 100 pareto optimal solutions. Moreover, the average value of the total mass of the robot is 3.9 kg.

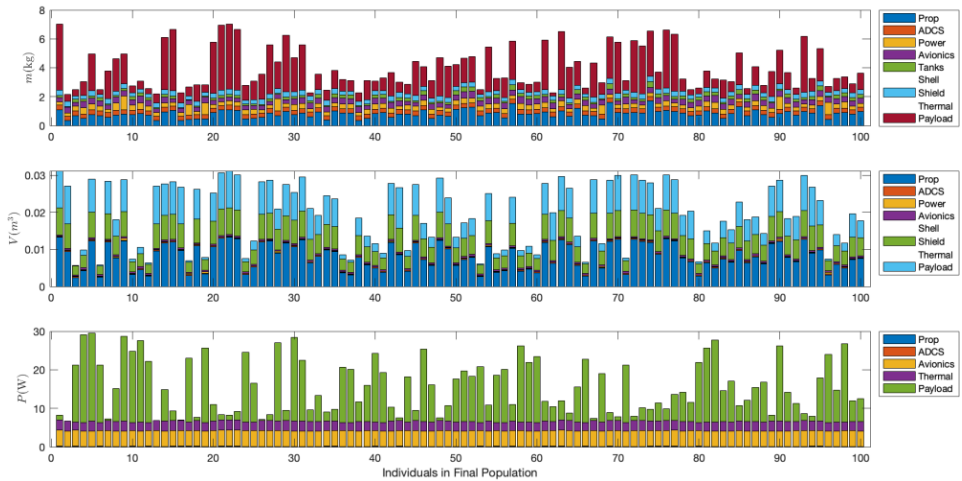

Fig. 8. Mass, volume and power budget of the 100 individuals in the pareto optimal front.

Fig. 9 shows the number of instances, different modes of mobility and power system is selected over generations. It can be seen from Fig. 9(Top-Left) that among the three modes of mobility (hopping, rolling, and wheeled), hopping is the most efficient one as the other two are rejected within 24 generations. It is also clear from Fig. 9(Top-Right) that among three modes of hopping mobility (propulsive, mechanical and reaction-wheel), propulsive hopping is the most efficient one as the other two are rejected within 14 generations.

Also, among the three propellants used for propulsive hopping, steam-propulsion is rejected within 69 generations, while neither RP1/$H_2O_2$, nor $H_2/O_2$ propulsion is rejected as shown in Fig. 9(Bottom-Left). This shows that both the options are viable for this mission scenario. Moreover, among the two power systems (battery and fuel cell), neither got rejected making both options viable as shown in Fig. 9(Bottom-Right).

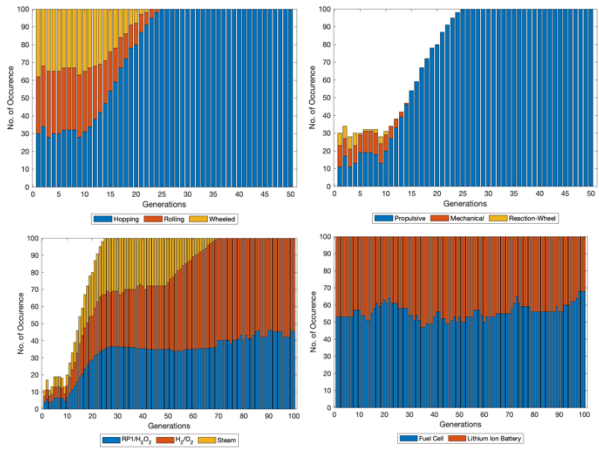

might extend to a few kilometres in length and so mission specification is to explore 1000m of the sublunarean void. The con-ops for performing this mission is shown in Fig. 10. A lander carrying multiple SphereX robots would descent nearby Mare Tranquilitatis Pit and deploy the robots one by one. Each robot will have three phases 1. Surface operation to approach the pit entrance, 2. Pit entrance maneuver, and 3. Sub-surface operation to explore the pit. The mission target is to explore 2000m on the surface in 10 hours, 50m in 10 minutes to enter the pit and 1000m inside the pit in 5 hours. The environmental conditions used for the simulations were gravity $g = 1.62\text{m/s}^2$, ambient temperature $T_a = 340\text{K}$ (surface), $T_a = 250\text{K}$ (sub-surface), radiation dose rate $I_0 = 100\text{rad/yr}$ (surface), $I_0 = 0\text{rad/yr}$ (sub-surface), and soil properties of Lunar soil. The constraints and the bounds on the design variables used were same as discussed in Section 5.1.

Fig. 9. (Top-Left) Number of instances hopping, rolling and wheeled modes of mobility selected over generations. (Top-Right) Number of instances propulsive, mechanical and reaction-wheel hopping modes of mobility selected over generations. (Bottom-Left) Number of instances $H_2/O_2$, $RP1/H_2O_2$ and steam based propulsive hopping modes of mobility selected over generations. (Bottom-Right) Number of instances fuel cells and lithium-ion batteries selected over generations.

## 5.2 Sub-surface exploration of Mare Tranquilitatis pit

The second simulation was run to perform sub-surface exploration of Mare Tranquilitatis Pit at 8.33°N 33.22°E. Lunar Reconnaissance Orbiter Camera (LROC) images reveal that the pit diameter ranges from 86 to 100m with a maximum depth from shadow measures of ~107m and that it opens into a sublunarean void of at least 20meters in extent. However, the sublunarean void

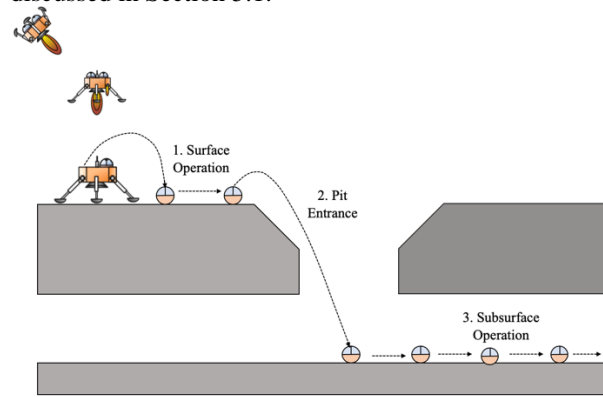

Fig. 10. Concepts of operation for exploring Lunar pits

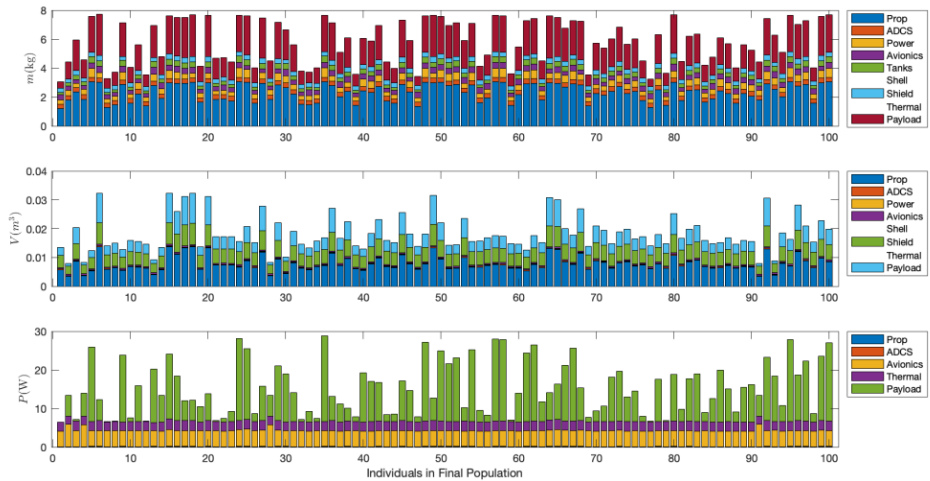

Fig. 11. Mass, volume and power budget of the 100 individuals in the pareto optimal front.

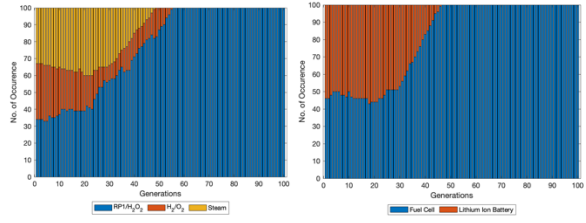

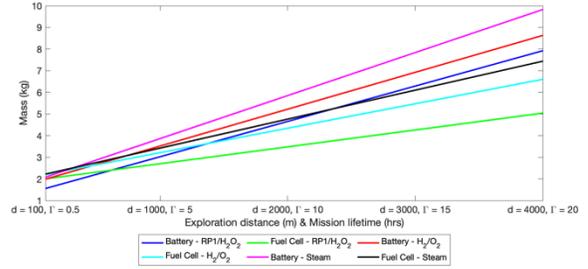

$$\min \mathbb{F}(\mathbb{x}) = m$$

$$subject \ to \begin{cases} \mathbb{G}_1(\mathbb{x}) \equiv (m - m_T)^2 = 0 \\ \mathbb{G}_2(\mathbb{x}) \equiv Index = 1 \end{cases} \quad (23)$$

Fig. 12. (Left) Number of instances $H_2/O_2$, $RP1/H_2O_2$ and steam based propulsive hopping modes of mobility selected over generations. (Right) Number of instances fuel cells and lithium-ion batteries selected over generations.

Fig. 11 shows the mass, volume, and power budget of the robot for the pareto-optimal solutions found. It can be seen that the minimum and maximum values of mass, volume and power available for the payload are $m_{pay}^{(L)} = 0.20$, $m_{pay}^{(U)} = 2.93 \ kg$ , $\mathcal{V}_{pay}^{(L)} = 0.00021$, $\mathcal{V}_{pay}^{(U)} = 0.0104 \ m^3$ , and $P_{pay}^{(L)} = 0.04$, $P_{pay}^{(U)} = 22.20 \ W$ . The average values of the mass, volume and power available for the payload are $\bar{m}_{pay} = 1.82 \ kg$ , $\bar{\mathcal{V}}_{pay} = 0.0047 \ m^3$ , and $\bar{P}_{pay} = 8.6 \ W$ over the 100 pareto optimal solutions. Moreover, it can be seen that average value of the total mass of the robot increased to 5.8 kg from 3.9 kg in test scenario 1. Fig. 12 shows the number of instances, different modes of mobility and power system is selected over generations. Also, among the three propellants used for propulsive hopping, steam-propulsion is rejected within 48 generations, while $H_2/O_2$ propulsion is rejected within 55 generations. This shows that $RP1/H_2O_2$ propulsion is the fittest mobility option for this mission scenario. Moreover, among the two power systems, battery system got rejected within 46 generations, thus making fuel cell power system as the fittest option. From Fig. 9 and 12 it is clear that the selection of mobility and power system depends on the mission exploration and mission time goals.

*5.2 Comparative Analysis*

Since, the selection of the propulsive hopping mobility and power system varied across the two mission scenarios presented in Section 5.1 and 5.2, a comparative study of the two systems is done for varying exploration distance and mission time. For the comparative study, the choice of the avionics was fixed, and the available payload mass, volume and power were considered 1kg, 10cm3, and 10W respectively. As such the problem is expressed as a single-objective optimization problem to minimize the mass of the robot with 2 design variables $\mathbb{x} = [m, r]$. For each design variable, the mass of each subsystem is calculated and then added together to find the total mass of the system $m_T$. Two constraints were added such that $m = m_T$, and the assembly index $Index = 1$. The optimization problem is mathematically formulated as Eq. (23).

Figure 13: Mass of the robot for all combinations of propulsive mobility system and power system for varying exploration distance and mission time.

Multiple simulations were performed for each combination of the propulsive hopping mobility and power system to find the optimal mass of the robot for varying exploration distance and mission time on the surface of the Moon. Fig. 13 shows the mass of the robot for each combination. It can be seen that for an exploration objective of $d_{targ} = 100 \ m$ and $\Gamma = 0.5 \ hrs$ , the system with lithium-ion batteries and $RP1/H_2O_2$ propulsive mobility is the optimal choice, however as the exploration objective increases to $d_{targ} = 4000 \ m$ and $\Gamma = 20 \ hrs$, the system with fuel cells and $RP1/H_2O_2$ propulsive mobility is the optimal choice. It can also be seen that for each of the propulsive mobility system, the ones with battery system is better than the ones with fuel cells for low exploration objectives, but as the exploration objectives increases the ones with fuel cells are far better.

## 6. Conclusions

The paper formulated and solved a multidisciplinary optimization (MDO) problem for SphereX, which included geometric design along with mobility and temperature control for planetary surface exploration missions. The problem was constructed with seven disciplines: mobility system, power system, communication, avionics, thermal, radiation shielding and shell which interacted with a COTS inventory for electronics, mobility controller for exploration and a thermal controller for maintaining the body temperature of the robot to find optimal design solutions for a specific planetary exploration mission. To solve the problem, the AMDCO framework was implemented that used a genetic algorithm based multi-objective optimizer at the system level to find the Pareto-optimal results while using gradient-based optimization techniques at the subsystem level. We have demonstrated that finding the optimal design variables associated with all the major disciplines of SphereX for a predefined exploration task is feasible through a rigorous multidisciplinary approach. The approach provides a system-level perspective of the problem with sufficient depth to capture high-level trade-offs and reveal insights that are perhaps not obvious at the discipline level. The solution provides a geometric solution that is useful for ground development of SphereX taking into consideration its operational and exploration goals on a target environment.

For operational point of view, each of the designs identified by the multidisciplinary optimization process needs further research and development. Future work will involve using these design solutions to perform path-planning with multiple robots to explore a target environment. In addition to that, hardware experimental results will be shown for exploring unknown environments like caves and lava tubes for mapping and localization.